%% file: IGLU-main.tex
\documentclass{article}

\usepackage[preprint]{neurips_2022}

\usepackage[utf8]{inputenc} % allow utf-8 input
\usepackage[T1]{fontenc}    % use 8-bit T1 fonts
\usepackage{hyperref}       % hyperlinks
\usepackage{url}            % simple URL typesetting
\usepackage{booktabs}       % professional-quality tables
\usepackage{amsfonts}       % blackboard math symbols
\usepackage{nicefrac}       % compact symbols for 1/2, etc.
\usepackage{microtype}      % microtypography
\usepackage{xcolor}         % colors
\usepackage{ulem}
\usepackage{amsmath}
\usepackage{amssymb}
\usepackage{xspace}
\usepackage{graphicx} % more modern
\usepackage[ruled,boxed,lined]{algorithm2e}
\usepackage{xcolor}
\usepackage{svg}
\usepackage[inline]{enumitem}
\usepackage{wrapfig}
\usepackage{listings}

\newcommand{\name}{IGLU\xspace}

\newcommand*\samethanks[1][\value{footnote}]{\footnotemark[#1]}

\newboolean{showcomments}
\setboolean{showcomments}{true}
\ifthenelse{\boolean{showcomments}}
  {\newcommand{\nb}[3]{
  {\color{#2}\small\fbox{\bfseries\sffamily\scriptsize#1}}
  {\color{#2}\sffamily\small$\triangleright~$\textit{\small #3}$~\triangleleft$}
  }
  }
  {\newcommand{\nb}[3]{}
  }

\colorlet{punct}{red!60!black}
\definecolor{background}{HTML}{EEEEEE}
\definecolor{delim}{RGB}{20,105,176}
\colorlet{numb}{magenta!60!black}
\lstdefinelanguage{json}{
    basicstyle=\normalfont\ttfamily,
    numbers=left,
    numberstyle=\scriptsize,
    stepnumber=1,
    numbersep=8pt,
    showstringspaces=false,
    breaklines=true,
    frame=lines,
    backgroundcolor=\color{background},
    literate=
     *{0}{{{\color{numb}0}}}{1}
      {1}{{{\color{numb}1}}}{1}
      {2}{{{\color{numb}2}}}{1}
      {3}{{{\color{numb}3}}}{1}
      {4}{{{\color{numb}4}}}{1}
      {5}{{{\color{numb}5}}}{1}
      {6}{{{\color{numb}6}}}{1}
      {7}{{{\color{numb}7}}}{1}
      {8}{{{\color{numb}8}}}{1}
      {9}{{{\color{numb}9}}}{1}
      {:}{{{\color{punct}{:}}}}{1}
      {,}{{{\color{punct}{,}}}}{1}
      {\{}{{{\color{delim}{\{}}}}{1}
      {\}}{{{\color{delim}{\}}}}}{1}
      {[}{{{\color{delim}{[}}}}{1}
      {]}{{{\color{delim}{]}}}}{1},
}

\title{IGLU 2022: Interactive Grounded Language Understanding \\ in a Collaborative Environment at NeurIPS 2022}
\author{
     Julia Kiseleva \thanks{Leader Organizers} \and
     Alexey Skrynnik\samethanks \and 
     Artem Zholus\samethanks \and
     Shrestha Mohanty \samethanks \and 
     Negar Arabzadeh \samethanks \and
     Marc-Alexandre Côté \samethanks \and
     Mohammad Aliannejadi\and
     Milagro Teruel \and
     Ziming Li \and  
     Mikhail Burtsev\and
     Maartje ter Hoeve\and
     Zoya Volovikova \and
     Aleksandr Panov\and
     Yuxuan Sun\and
     Kavya Srinet\and 
     Arthur Szlam\and
     Ahmed Awadallah \\
{\tt julia.kiseleva@microsoft.com}} 
\begin{document}

\maketitle

\begin{abstract}
Human intelligence has the remarkable ability to quickly adapt to new tasks and environments. Starting from a very young age, humans acquire new skills and learn how to solve new tasks either by imitating the behavior of others or by following provided natural language instructions. To facilitate research in this direction, we propose \textit{IGLU: Interactive Grounded Language Understanding in a Collaborative Environment}.
The primary goal of the competition is to approach the problem of how to develop interactive embodied agents that learn to solve a task while provided with grounded natural language instructions in a collaborative environment. Understanding the complexity of the challenge, we split it into sub-tasks to make it feasible for participants.  
This research challenge is naturally related, but not limited, to two fields of study that are highly relevant to the NeurIPS community: Natural Language Understanding and Generation (NLU/G) and Reinforcement Learning (RL). Therefore, the suggested challenge can bring two communities together to approach one of the important challenges in AI. Another important aspect of the challenge is the dedication to perform a human-in-the-loop evaluation as a final evaluation for the agents developed by contestants.
\end{abstract}

\subsection*{Keywords}
%Up to five, from generic to specific.
Natural Language Understanding (NLU), Reinforcement Learning (RL), Grounded Learning, Interactive Learning, Embodied RL.

\input{01-desc}
\input{03-resources}

\bibliographystyle{plainnat}
\bibliography{ref}
\appendix
\input{04-appendix}

\end{document}

%% file: 01-desc.tex
\section{Competition Description}

\subsection{Background and Impact}

%TODO: Provide some background on the problem approached by the competition and fields of research involved. A good outcome of a competition is to learn something new by answering a scientific question or make a significant technical advance.  Describe the scope and indicate the anticipated impact of the {proposed} competition (economical, humanitarian, societal, etc.).

Humans have the remarkable ability to adapt to new tasks and environments quickly. Starting from a very young age, humans acquire new skills and learn how to solve new tasks either by imitating the behavior of others or by following provided natural language instructions~\citep{an_imitation_1988, council_how_1999}. Studies in developmental psychology have shown evidence of natural language communication being an effective method for transmission of generic knowledge between individuals as young as infants~\citep{csibra2009natural}. This form of learning can even accelerate the acquisition of new skills by avoiding trial-and-error when learning only from observations~\citep{thomaz2019interaction}. 
Inspired by this, the AI research community attempts to develop grounded interactive \textit{}{embodied agents} that are capable of engaging in natural back-and-forth dialog with humans to assist them in completing real-world tasks~\citep{winograd1971procedures,narayan2017towards, levinson2019tom,chen2020ask,abramson2020imitating,arabzadeh2022preme,li2021improving,li2020rethinking}. Notably, the agent needs to understand when to initiate feedback requests if communication fails or instructions are not clear and requires learning new domain-specific vocabulary~\citep{Aliannejadi_convAI3, aliannejadi2021building,rao2018learning, narayan2019collaborative, jayannavar-etal-2020-learning}.
Despite all these efforts, the task is far from solved.
For that reason, we propose the \name competition, which stands for Interactive Grounded Language Understanding (IGLU) in a collaborative environment.

%Justify the relevance of the problem to the NeurIPS community and estimate the number of participants in your proposed competition: will your competition will be of interest to a large audience or limited to a small number of domain experts? 
Specifically, the goal of our competition is to approach the following scientific challenge: 
\textit{How to build interactive embodied agents that learn to solve a task while provided with grounded natural language instructions in a collaborative environment?}
By \textit{`interactive agent'} we mean that the agent can: (1) follow the instructions correctly, (2) ask for clarification when needed, and (3) quickly adapt newly acquired skills. The \name challenge is naturally related to two fields of study that are highly relevant to the NeurIPS community: Natural Language Understanding and Generation (NLU / NLG) and Reinforcement Learning (RL).

\noindent
\paragraph{Relevance of NLU/G} 
Natural language interfaces (NLIs) have been the ``holy grail'' of human-computer interaction for decades~\citep{woods1972lunar, codd1974seven, hendrix1978developing}. The recent advances in language understanding capability~\citep{devlin2018bert, LiuRoberta_2019, clark2020electra, adiwardana2020towards, roller2020recipes, brown2020language, ouyang2022training} powered by large-scale deep learning has led to a major resurgence of natural language interfaces such as virtual assistants, dialog systems, semantic parsing, and question answering systems etc~\citep{liu2017iterative, liu2018adversarial, dinan2020second, zhang2019dialogpt}.
%The horizon of NLIs has also been significantly  expanding  from databases~\citep{copestake1990natural} to,  knowledge  bases~\citep{berant2013semantic}, robots~\citep{tellex2011understanding}, Internet of Things (virtual  assistants like  Siri  and Alexa), Web service APIs~\citep{su2017building}, and other forms of interaction~\citep{fast2018iris, desai2016program, young2013pomdp}.
Recent efforts have also focused on interactivity and continuous learning to enable agents to interact with users to resolve the knowledge gap between them for better accuracy and transparency. This includes systems that can learn new task from instructions~\citep{li-etal-2020-interactive}, assess their uncertainty~\citep{yao-etal-2019-model}, ask clarifying questions~\citep{aliannejadi2021building} and seek and leverage feedback from humans to correct mistakes ~\citep{elgohary-etal-2020-speak}.

\noindent
\paragraph{Relevance of RL} Recently developed RL methods celebrated successes for a number of tasks~\citep{bellemare2013arcade, mnih2015human, mnih2016asynchronous, silver2017mastering, hessel2018rainbow, paul2021learning}. One of the aspects that helped to speed up RL methods development is game-based environments, which provide clear goals for an agent to achieve in flexible training settings. However, training RL agents that can follow human instructions has attracted fewer exploration~\citep{chevalier2019babyai,cideron2019self,hu2019hierarchical, chen2020ask, shu2017hierarchical}, due to complexity of the task and lack of proper experimental environments. Another interesting aspect of language in the context of language conditioned RL is its compositionality. Conditioning on language abstractions might help the agent to better understand visually different but semantically similar situations~\citep{mu22abstractions}.
~\citet{shu2017hierarchical} proposed a hierarchical policy modulated by a stochastic temporal grammar for efficient multi-task reinforcement learning where each learned task corresponds to a human language description in Minecraft environment. The BabyAI platform~\citep{chevalier2019babyai} aims to support investigations towards learning to perform language instructions with a simulated human in the loop. ~\citet{chen2020ask} demonstrated that using step-by-step human demonstrations in the form of natural language instructions and action trajectories can facilitate the decomposition of complex tasks in a crafting environment. 

\paragraph{Relevance to Real Live Scenarios and Societal Impact}
Our competition can drive positive impact for real-life scenarios such as:
\begin{itemize}[nosep, leftmargin=*]
    \item \textbf{Education:} 
    \textit{Minecraft: Education Edition}\footnote{\url{https://education.minecraft.net/}} is a game-based learning platform that promotes creativity, collaboration, and problem-solving in an immersive digital environment. As of 2021, educators in more than $115$ countries are using Minecraft across the curriculum. As stated in~\citet{url-minecraft-edu}, adding AI elements to this educational platform will move its potential to a new level. AI applications have the power to become a great equalizer in education. Students can get personalized education and scaffolding while being less dependent on uncontrollable factors such as the quality of their teachers or the amount of help they receive from their caregivers.
    %\item \textbf{Gaming:} Game designers could leverage this technology to develop richer NPCs (non-player characters) for the players to interact with. Players could also benefit from personalized quests in open world games. 
    \item \textbf{Robotics:} \citet{bisk-etal-2016-natural} proposed a protocol and framework for collecting data on human-robot interaction through natural language. The work demonstrated the potential for unrestricted contextually grounded communications between human and robots in blocks world. Developing robots to assist humans in different tasks at home has attracted much attention in the Robotics field~\citep{stuckler2012robocup}. In fact, the Robocup@Home\footnote{\url{https://athome.robocup.org/}} and the Room-Across-Room\footnote{\url{https://ai.google.com/research/rxr/habitat}} have run for several years. Given that the main human-robot interaction is through dialog, and the robot is supposed to assist the human in multiple tasks, we envision \name to enable more effective task grounded dialog training between human and robots.
\end{itemize}

\subsection{Novelty}
%Indicate whether this is an entirely new competition, or part of a series, eventually re-using old data. 
%If this is an updated version of a previously accepted competition (either at NeurIPS or another venue), extensively comment on the updates you introduced and motivate if there are not. 
%If you are aware of similar previous competitions, describe the key differences with your proposal.

Last year (@NeurIPS2021), we organized the first edition of the IGLU competition to tackle the task of grounded language understanding and interactive learning and bring together the NLU/G and RL research communities~\cite{kiseleva2021neurips,kiseleva2022interactive}. In light of the competition results, it is clear the task is far from being solved and requires more collective effort to explore promising directions. Hence, we now proposed the second edition of the IGLU competition.

Learning from the feedback we receives and from our own experiences running the IGLU in 2021, we made some significant changes for this year's edition. This year our main focus is on \textit{interactive agents}. To accommodate it, we made the following changes:
\begin{itemize}[nosep, leftmargin=*]
\item \textit{Simplifying data collection}: in order to increase the speed of the data collection and, subsequently, collect more data, we employ a single-turn data collection strategy instead of a multi-turn one (Section~\ref{data collection}).
\item \textit{Speeding up training environment}: the critical change is the use of a new gridworld environment that allows for fast and scalable turnaround experiments while requiring a lower amount of memory and computation resources (Section \ref{sec:code_rl})
\item \textit{Reformulating NLP task:} the NLP task is reformulated to tackle the generation of clarifying question that leads to truly interactive agents (Section \ref{sect:clarifying_questions}).
\end{itemize}

\subsection{Data}
\label{sec:data}
\name is partially motivated by the HCRC Map Task Corpus~\citep{thompson_hcrc_1993}, which consists of route-following multi-turn dialogs between an \textit{Instruction Giver} (further called Architect) and a \textit{Follower} (further called Builder).
To enable data collection for \name we have developed a tool by leveraging the CraftAssist library, which can be easily plugged into crowdsourcing platforms, to collect the dataset of multi-turn interactions between Architect and Builder to extend the original one~\cite{jayannavar-etal-2020-learning}. \href{https://youtu.be/Ls6Wv7EUDA0}{Here}\footnote{\url{https://youtu.be/Ls6Wv7EUDA0}} is a visualization of the collected interactions. \footnote{The developed tool for data collection is used for \name 2022}.

We modified the data collection strategy based on the lessons learned from last year's contest. To make the contest appealing to a broader audience, we simplify the task by leveraging the data we collected last year. Namely, we focus on the setup where an interactive agent is dropped in the middle of an ongoing game where the structure is built partially (Fig.~\ref{fig:single-turn-example}). Then, the agent is given a natural language instruction to be executed in the current world, and in case of uncertainty, they need to ask clarifying questions. Since the current data collection is limited to building-related tasks, the agent and builder are interchangeably used. We are working on enhancing the data collecting tool to allow agents to perform various tasks, such as `grab,' `bring,' etc., to convert a builder into an actor.

\subsubsection{Data Collection}
\label{data collection}
\begin{wrapfigure}{r}{0.5\textwidth}
 \begin{center}
   \vspace{-0.5cm}
   \includegraphics[clip, width=0.5\textwidth]{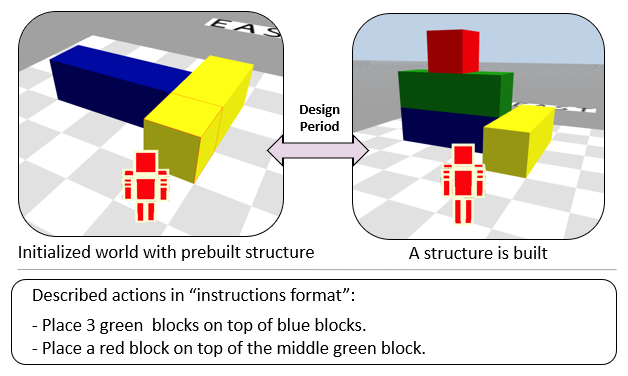}
   \caption{An example of data collection}
   \vspace{-0.5cm}
   \label{fig:single-turn-example}
\end{center}  
\end{wrapfigure}

To extend the dataset collected during \name 2021\footnote{The previously agreed IRB is reused}, we will follow a simplified procedure by requiring only one player per game (instead of two). The data collection task is as follows:
\begin{enumerate} [nosep, leftmargin=*]
    \item a player is dropped in the middle of an ongoing game, and they need to extend the structure in an imaginary way (Fig.~\ref{fig:single-turn-example}: Left) within a certain time limit; 
    \item once the structure the player had in mind is done, they need to write down instructions on how to remodel what they just constructed. The instructions should be descriptive and sufficient as if they wanted to instruct someone else to redo what they have done (Fig.~\ref{fig:single-turn-example}: Right). 
\end{enumerate}

We extend the collected single-turn dataset with clarifying questions by initializing the second data collection stage. The collected instructions are shown to players, and they are asked either to execute the instruction or ask clarifying questions when needed.
The multi-turn data contains dialogs between the Architect and the Builder, the single-turn data contains the retrospective descriptions of what the Builder has done. Both types of data contains pairs of starting and ending gridworlds (per-turn). And, importantly, we also collect streams of an embodied behavior which were captured from human executing single or multi-turn tasks. These streams of events can be converted to the environment demonstration data using the data adaptation tool\footnote{\href{https://github.com/iglu-contest/iglu/blob/master/iglu/data/}{https://github.com/iglu-contest/iglu/blob/master/iglu/data/}} that we provide. We envision this data to be used in the role of the training data for imitation learning or reinforcement learning agents. An example of a behavior stream logs is present in Appendix~\ref{behavior_data}.

% If your competition evaluates submissions based on the analysis of data, please provide detailed information about the availability of the evaluation data and their annotations, as well as permissions or licenses to use such data.

% If new data were collected or generated for the purpose of the competition, provide details on the procedure, including permissions to collect such data obtained by an ethics committee, if human subjects are involved. In this case, it must be clear in the document that the data will be ready and approved for use prior to the official launch of the competition. 

% \textbf{Please justify that:} 1) you have access to large
% enough data-sets to make the competition interesting and draw
% conclusive and statistically significant results; 2) the data will be made freely available for the competition; 3) the ground truth has has not been previously published and has been kept confidential.

\subsection{Tasks and Application Scenarios}
\begin{wrapfigure}{r}{0.5\textwidth}
 \begin{center}
   \vspace{-1.5cm}
   \includegraphics[width=0.5\textwidth]{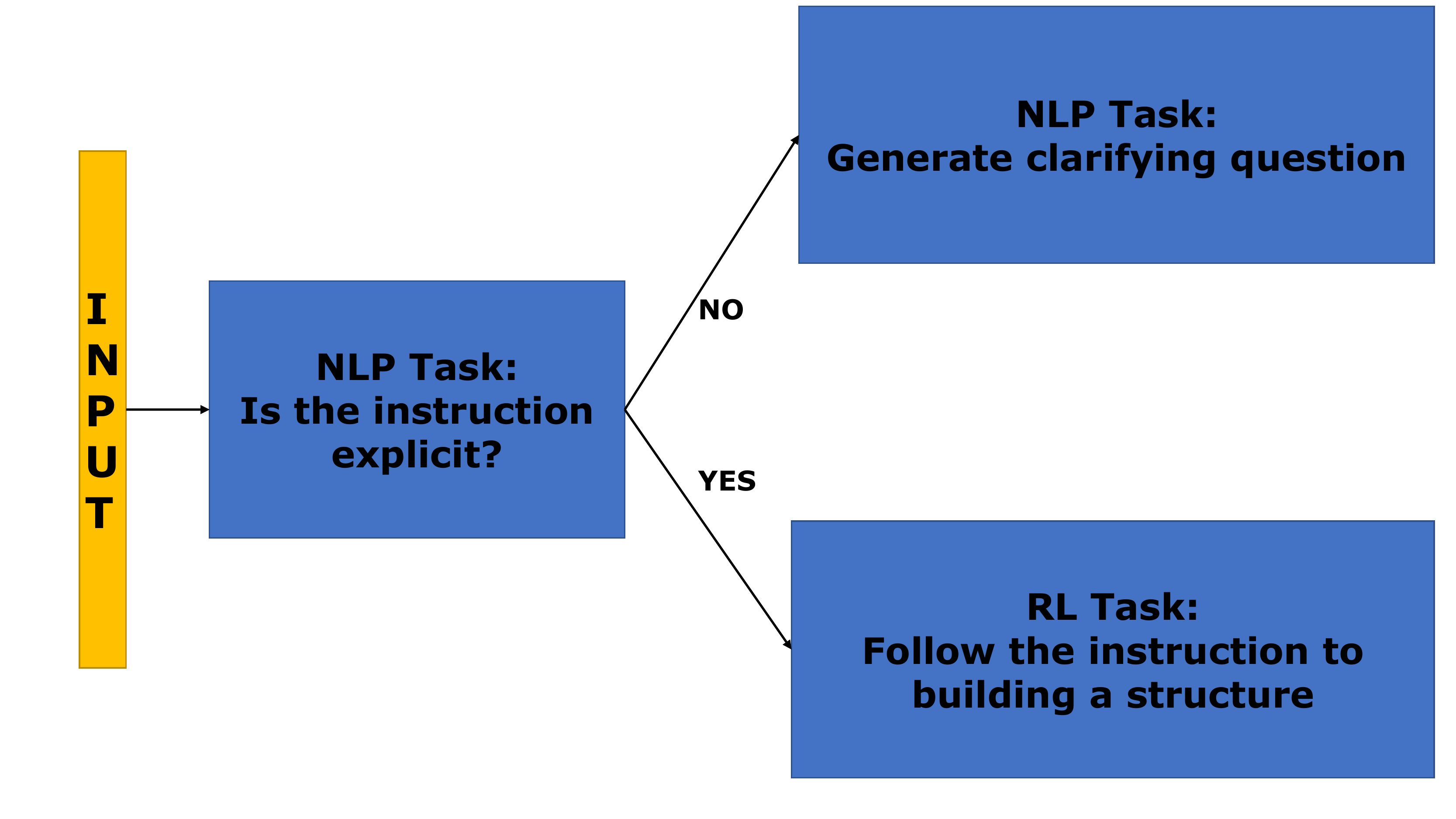}
   \caption{IGLU's general overview}
   \end{center}
   \vspace{-0.5 cm}
   \label{fig:iglu}
\end{wrapfigure}

The considered challenge of developing an \textit{`interactive agent'} naturally splits into two main tasks visualized in Fig.~\ref{fig:iglu} suggests: 
\begin{itemize}[nosep, leftmargin=*]
\item NLP-related: to decide if the provided grounded instruction needs to be clarified and generate clarifying question if it's a case;
\item IR-related: to follow a clear grounded instruction in order to complete the described task.
\end{itemize}
Our vision suggests that the successful solutions to presented tasks can be combined into the end-to-end pipeline to develop desired interactive agents.

\subsubsection{NLP Task: Asking Clarifying Questions}
\label{sect:clarifying_questions}
%We hope to attract participants from NLP community to predict if the input from the architect is clear and sufficient to rebuild the structure. In case the builder is given a vague instruction, it is required to issue a question to obtain more details on the instruction.
Inspired by~\cite{aliannejadi2021building}, we split the problem into the following concrete research questions:

\begin{enumerate}[label=\textbf{RQ\arabic*},nosep,leftmargin=*]
 \item {When to ask clarifying questions?} \\
 Given the instruction from the Architect, a model needs to predict whether that instruction is sufficient to complete the described task or further clarification is needed. Simply put, here it is decided if we need to activate the Builder.
 \item {What clarifying question to ask?} \\
  If the given instruction from the Architect is ambiguous, a clarifying question should be raised. In this research question, we are specifically interested in "what to ask" to clarify the given instruction. The original instruction and its clarification can be used as an input for the Builder.
\end{enumerate}

As a starting baseline we use implementation provided in~\cite{aliannejadi2021building}.

%For this task, we collect another round of data by reversing the performed task in section \ref{data collection}. Given the initialized world and the collected instructions, we ask annotators if the provided instructions are ambiguous or clear enough. If annotators diagnose that the instructions can be interpreted in multiple ways and require more details, we ask them to issue clarifying questions in order to to make the instructions clear.  We further ask them to construct the given instructions in the initialized Voxel world and compare it to the target structure that the initial architect built. Therefore, we can observe if the instructions were truly clear enough to remodel the task or not.

% Describe the competition's task(s) and explain to which specific real-world scenario(s) they correspond. Put particular emphasis on how the competition relates to a real problem faced in industry or academia. If it cannot be cast in those terms, provide a detailed hypothetical application scenario and focus on its relevance to NeurIPS.

% Justify that the problem posed are scientifically or technically challenging but not impossible to solve. If data are used, consider illustrating the same scientific problem using several data-sets from other application domains.

\subsubsection{RL Task: Building Structures}
Our experience of \name 2021 emphasized the complication of the builder task. Indeed, the optimal builder needs to know how to build any structure. The state-space of the possible cubes combinations is very large: (11, 11, 9) places where an agent can build a cube with six different colors. We believe that to solve such a challenging task, one needs a large-scale training procedure and the possibility to run a spectrum of various experiments. Unfortunately, the last part was restricted by the slow sampling rate of the Malmo environment~\cite{kanervisto2022}. Thus, for \name 2022 we worked in two directions: speeding up the environment and creating a baseline, which can be generalized to build various structures.

\subsection{Baselines, Code, and Material Provided}
% Specify what the baseline solution(s) for the competition will be and provide preliminary results.

% Indicate \textbf{how} and \textbf{when} you plan to release the \textit{``starting kit''} for your competition. This should include code for baselines and data loading tools to help the participants easily join the competition. For certain competitions, material provided may include a hardware platform.

\subsubsection{IGLU Gridworld RL Environment}
\label{sec:code_rl}
For the builder task, we created a novel RL environment\footnote{\href{https://github.com/iglu-contest/gridworld}{https://github.com/iglu-contest/gridworld}} where an embodied agent can build spatial structures of blocks of different colors. The agent's goal is to complete a task expressed as an instruction written in natural language.

\begin{wrapfigure}{r}{.45\textwidth}
 \centering
 \vspace{-20pt}
 \includegraphics[width=\linewidth]{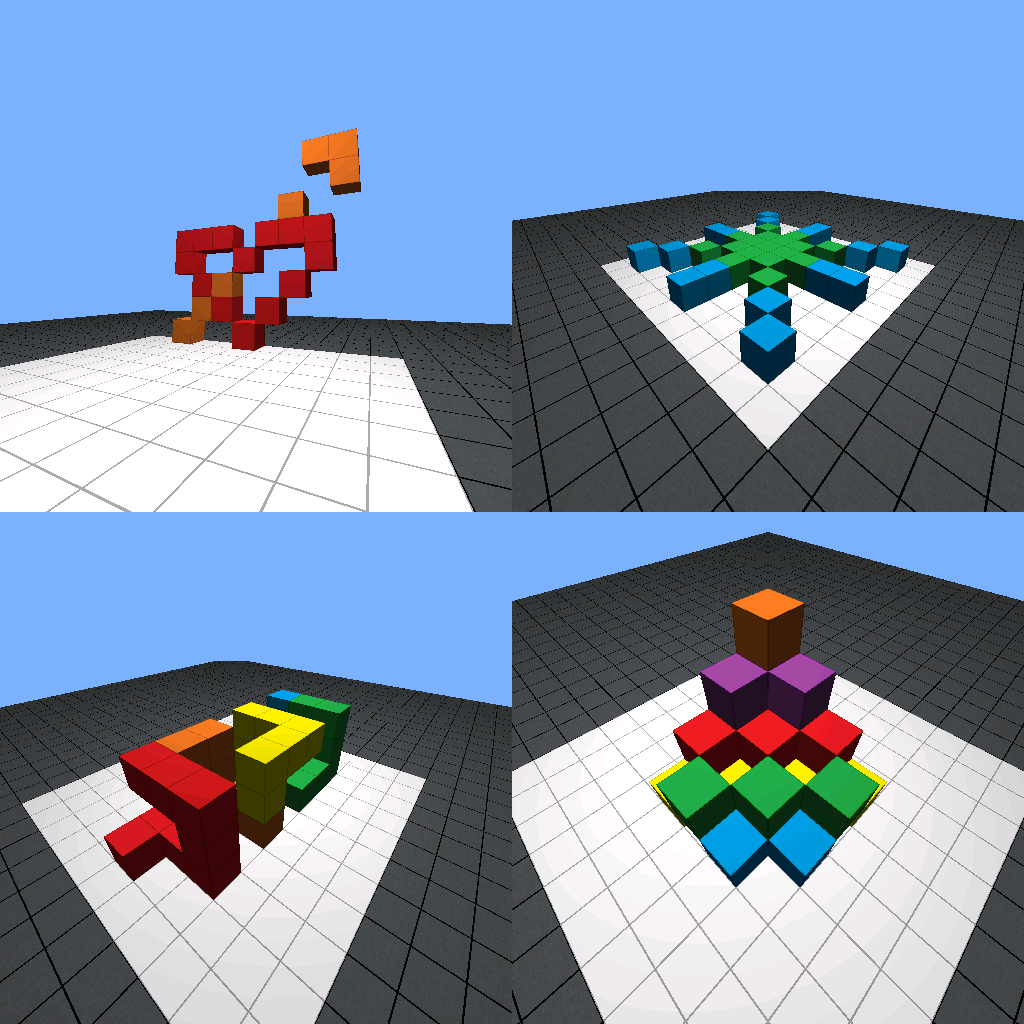}
 \caption{Examples of target structures rendered by the gridworld environment}
 \vspace{-10pt}
\end{wrapfigure}

We implement the environment in pure Python using a simplified version of an open-source Minecraft clone\footnote{\href{https://github.com/fogleman/Minecraft}{https://github.com/fogleman/Minecraft}}. The renderer is decoupled from the core environment, it can run headless with GPU-acceleration, and can be disabled. The simplicity of the environment makes it fast and scalable for highly loaded RL experiments. Without a renderer, the environment runs at 17000 steps-per-second (SPS) and at 4000 SPS with rendering enabled,  which is $85$ (for vector observations) and $20$ (for image-based observations) times faster than the previous version of the \name@2021 environment. Such significant speedup is possible by removing the complexity of using Minecraft, which comes with a massive amount of (for now) unused features. The latest version, in turn, is as simple as it is possible, that is why it is much faster.

The observation space consists of
point-of-view (POV) image $(64, 64, 3)$, inventory item counts $(6,)$, a snapshot of the building zone $(11, 11, 9)$, and the agent's $(X, Y, Z)$ position with pitch and yaw angles $(5,)$. The building zone is represented via a 3D tensor with block ids (e.g., $0$ for air, $1$ for blue block, etc.). The agent can navigate over the building zone, place and break blocks, and switch between block types.
Additionally, the environment provides a dialog from the dataset, which defines a task for building. The dialog is split into two parts: the first one is a context for the task, and the second one defines the target. The context utterances define blocks that are placed before the beginning of the episode. We thus initialize the episode with such blocks placed at their correct positions. Target utterances define the rest of the blocks needed to be added. The action space combines discrete and continuous subspaces:  
a discrete space of 14 actions: \textit{noop, step forward, step backward, step right, step left, jump, brake block, place block, choose block type 1-6} and two-dimensional continuous camera movement space. 

The reward in the environment reflects how complete the built structure is agnostic to its location. We run a convolution-like procedure between the state grid and the target grid. We calculate the per-block intersection for each alignment and then maximize it over alignments. The pseudocode for the reward calculation is presented in Appendix~\ref{reward_calc}.

\subsubsection{Builder Baseline}
\label{sec:baselines}

Training an agent to build any user-defined structure is a challenging task. Even if the task given to the agent is not described in natural language (e.g., by providing the target structure of the grid), it is still nontrivial. Thus, in addition to last year's single-task baselines\footnote{\href{https://github.com/iglu-contest/iglu-builder-baseline-dreamer}{https://github.com/iglu-contest/iglu-builder-baseline-dreamer}}\footnote{\href{https://github.com/iglu-contest/iglu-builder-baseline-rllib}{https://github.com/iglu-contest/iglu-builder-baseline-rllib}}, we are providing a multi-task one. We focus on an approach that can solve multiple problems for that new baseline, including problems not seen during the training phase. The whole builder problem can be divided into two main parts, which could be solved by NLU and RL methods (see Fig.~\ref{fig:baseline-decomposition} in Appendix). This section provides an approach to building structure given a target grid. 

\begin{figure}[ht!]
\centering
\includegraphics[ width=0.98\textwidth]{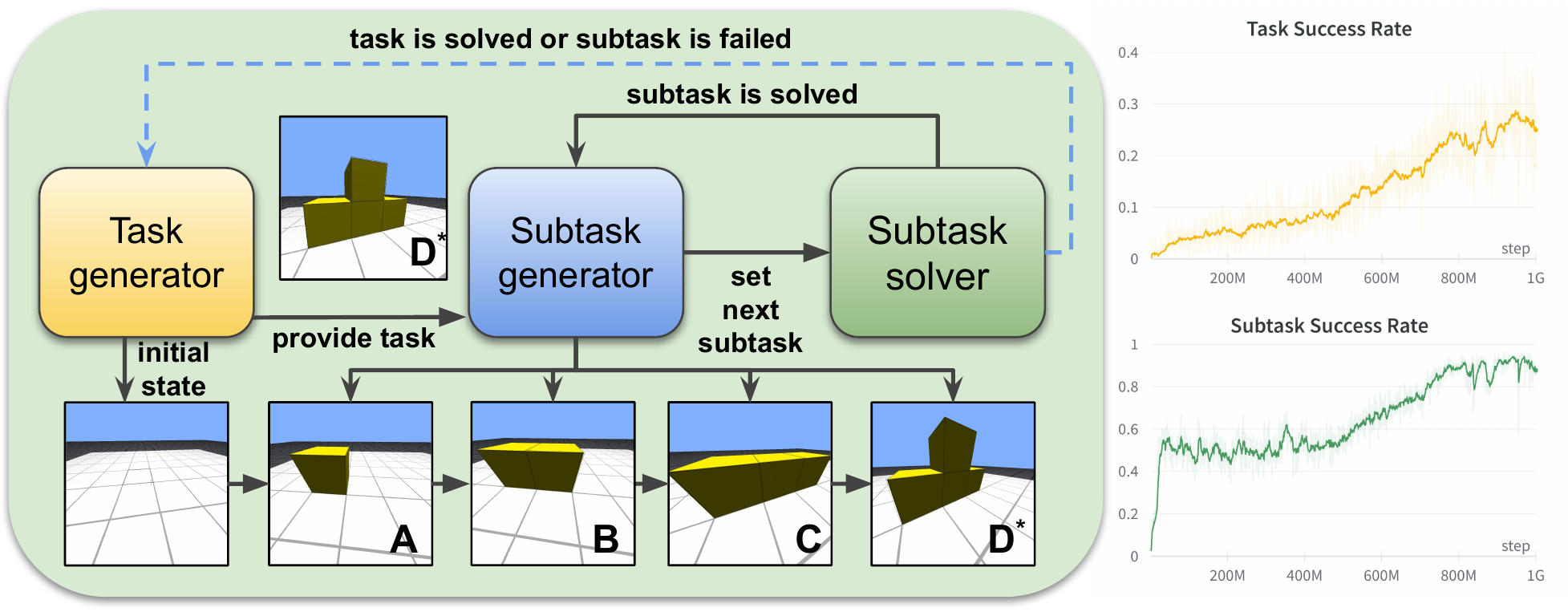}
   \caption{Scheme of the multi-task research baseline. The agent is trained to build any structure using a Task generator and heuristic approach, which determines the following block to place for the agent (Subtask generator). The episode is restricted to adding or removing only a single cube, which significantly simplifies the global task. Training curves for the PPO are presented on the right. The agent solves subtask in almost 90\% of episodes and solves the whole task in 25\% of trials.}
%   \vspace{-0.5cm}
   \label{fig:baseline}
\end{figure}

There are many ways to build a structure in \name environment. 
To overcome this, we split the whole system from the agents' perspective into three modules: a task generator module, a subtask generator module, and a subtask solving module (see left part of Fig.~\ref{fig:baseline}). We define the subtask as an episode of adding or removing a single cube. It allows us to train an agent with a dense reward signal in episodes with a short horizon. 

The right part of Fig.~\ref{fig:baseline} reports the learning curves of the PPO algorithm. The agent was trained for one billion steps using the Sample Factory framework~\cite{petrenko2020sample} on a single GPU machine in 15 hours. We use randomly generated compact structures as tasks and a subtask generating module. This module determines the order of blocks for subtasks. In this experiment, we used a heuristic one, which samples the next cube sequentially going through the 3D voxel of a target, but this module can be trained with a learning model. The baseline uses vector-based observation space (environment without rendering) and shows the training framerate of  19000 SPS. The PPO input is the agent's position, a compass angle, and target/current grid (3D voxels). 

To train the agent, we design a reward function, which defines the reward signal based on the Manhattan distance of the placed cube to the target position (see Table~\ref{table:reward-function}). In addition, the agent receives an extra reward, which is equal to $0.5$ if the agent puts the cube in the right place, under his feet. This addition sufficiently increased the agent's performance in tasks with high structures. 

% The full reward range was [1, 0.25, 0.05, 0.001, -0.0001, -0.001, -0.01, -0.02]. 
% For example , if the agent put the cube in the right place, he received a reward equal to 1; if he made a mistake by one cube, he received a reward equal to 0.25, etc. 
% if he made a mistake by two cubes - 0.05, etc. 
% For each mistake that was greater than seven blocks, he received a reward less than 0.01 of the previous values in the range. 

\begin{table}[ht!]
    \centering
    % \resizebox{\columnwidth}{!}{
    \caption{The design of the reward function is based on Manhattan distance. The agent would receive a negative reward if the distance of the placed cube from the target was more than three blocks. }
    \label{table:reward-function}
    
    \begin{tabular}{cccccccc}
    \toprule
    Distance & $0$ & $1$ & $2$ & $3$ & $4$ & $5$ & $i\in\{6, +\infty\}$ \\
    \midrule
    Reward & $1$ & $0.25$ & $0.05$ & $0.001$ & $-0.0001$ & $-0.001$ & $-0.01 \times (i-5)$  \\
    \bottomrule
    \end{tabular}
    % }
    
\end{table}

\subsection{Metrics}
\label{sec:metrics}

\paragraph{NLP Task} We evaluate this task in two distinct steps. The first step is a classification problem on whether we need clarification question or not. To compare the performance, we rely on precision, recall, $F_1$ score, and accuracy. \citet{aliannejadi2021building} used a similar approach for open-dialogue question-answering and noted that it is rather difficult and potentially can benefit from more exploration. For the second research question (i.e., "what to ask"), we evaluate how closely the generated questions resemble the human
issued clarifying questions. We adopt standard word-overlap metrics such as BLEU \cite{papineni2002bleu} score as well as word embedding based metrics that captures the contextualized and semantic similarity of the generated clarifying questions w.r.t the question issued by human annotators \cite{liu2016not}.

\paragraph{RL Task} For the Builder task, the main metric is $F_1$ score where the ``ground truth'' is the target structure (a 3d tensor) and the ``prediction'' is the snapshot of the building zone at the end of the episode (also a 3d tensor). This is different from \name 2021 where we made an additional maximization over timesteps of the metric. That maximization led to a bias towards specific types of behavior (e.g., uniformly random over actions). This year, we neglect this bias by reporting the metric at the last step of the episode. The episode terminates either when the structure is complete, or when the time limit is reached. The case when the episode ends at time limit is outside of the control of an agent. Therefore, we let the agent to decide when to end an episode (as a separate action). We report the performance by calculating the intersection between the last step snapshot of the building zone and the target zone. This value is then used to calculate the $F_1$ score.

Additionally, to provide more information to participants, we provide a skill labeling of tasks. Each label indicate a specific skill that the agent should possess in order to build a particular structure. For example, a structure that has blocks (or groups of block) that fly in the air without touching any other blocks side-by-side. This kind of structures requires to understand that one should place a supporting block and eventually remove it (as block placement is done side-by-side). We believe this information could help the participants to better understand their agents' capabilities. At the same time, the final metric is not related to the skills, it is just an average taken over all tasks.
The whole set of highlighted skills is present in Appendix \ref{skills}.
% For quantitative evaluations, select one or more scoring metrics and justify that they effectively assess the efficacy of solving the problem at hand. If no metrics are used, explain why and how the evaluation
% will be carried out. Explain how error bars will be computed and/or how the significance in performance difference between participants will be evaluated.

% You can include subjective measures provided by human judges. In that case, describe the judging criteria, which must be as orthogonal as possible, sensible, and specific. Provide details on the judging protocol, especially how to break ties between judges. Explain how judges will be recruited and, if possible, give a tentative list of judges, justifying their qualifications.  

\subsection{Website, Tutorial and Documentation}

Competition website is hosted at \href{https://www.iglu-contest.net/}{iglu-contest.net}. The main website contains links to all materials such as the competition slack channel, documentation for the RL environment\footnote{\href{https://iglu-contest.github.io/}{https://iglu-contest.github.io/}}, the Codalab competitions links. A dedicated email address for the competition is \href{mailto:info@iglu-contest.net}{info@iglu-contest.net}. To ease reaching out the participants, we created a twitter account dedicated to the competition announcements\footnote{\href{https://twitter.com/IgluContest}{https://twitter.com/IgluContest}}.

% Include a link to the competition website or a tentative one if not ready yet. The website should be self-contained, presenting all the relevant information about the competition timeline and illustrating the necessary steps to participate. 

% It is strongly suggested to have a \textbf{FAQ/Tutorial section} and a \textbf{dedicated email address} to reach the organizers. These should be highlighted, easily reachable from the home page, made available from the start and regularly updated.
% The website must be online within two weeks after the acceptance notification. The website's entire content should be released as soon as possible and with enough advance on the start of the competition to allow the participants to have time to prepare.

% If available, provide a reference to a published paper or a white paper you wrote describing the problem, and/or explain what tutorial material you will provide.

%% file: 03-resources.tex
\section{Resources}

\subsection{Organizing Team}

Our team can be split into two main groups based on expertise: NLU/G and RL researchers.

\paragraph{\emph{The RL sub-team includes the following members:}}

\noindent
\paragraph{Artem Zholus} is an incoming summer research intern at EPFL and a Master's student at MIPT. His research focuses on model-based reinforcement learning with application to robotics. Artem has worked on many machine learning projects, such as reinforcement learning for drug discovery and generative models. In this project, he leads the RL environment development and contributes to the data collection, baselines development, and the competition infrastructure development.

\noindent
\paragraph{Zoya Volovikova} is a  Master's student at ITMO. Zoya worked on the development of scientific software using machine learning algorithms. As an example, a system for predicting the phase of the wavefront for an adaptive optical system. In the IGLU contest, she takes part in the development of RL baselines.

\noindent
\paragraph{Alexey Skrynnik} is an individual contributor. His current research focused on multi-agent, object oriented and hierarchical reinforcement learning. He is highly experienced in practical RL. Moreover, Alexey previously won MineRL Diamond Competition 2019 as a leader of the CDS team. In the IGLU contest, he leads the development of RL baselines.

\noindent
\paragraph{Aleksandr Panov} has managed the CDS team that won the MineRL Diamond Competition 2019. Currently, Aleksandr works in model-based RL and the use of RL methods in the task of navigation in indoor environments.

\noindent
\paragraph{Marc-Alexandre C\^ot\'e} is a Principal Researcher at Microsoft Research Montréal. His research interests lie at the intersection of reinforcement learning and language understanding. He leads the Microsoft TextWorld project which aims at developing new RL agents capable of navigating and interacting with text environments (e.g., text-based adventure games). Such agents should possess skills like reading and understanding natural language text, information/knowledge-gathering, planning, dealing with combinatorially large action-space. In terms of past experiences organizing competition, in 2019 Marc initiated the \href{https://aka.ms/ftwp}{First TextWorld Problem} competition presented at CoG 2019 as a way of promoting the TextWorld framework.

\noindent
\paragraph{\emph{The NLU/G sub-team consists of the following members:}}

\noindent
\paragraph{Julia Kiseleva} is a Senior Researcher at Microsoft Research. Her research interests are in natural language processing, information retrieval, and machine learning, with a strong focus on continuous learning from user feedback and interactions. Julia has been involved in many initiatives such as a serious of Search-Oriented Conversational AI (SCAI) workshop, which was continuously organized from 2015 - 2020. She also co-organized the recent challenge ConvAI3: Clarifying Questions for Open-Domain Dialogue Systems (ClariQ) at EMNLP2020.

\noindent
\paragraph{Negar Arabzadeh} is a Research Scientist at Spotify Research working on podcast retrieval evaluation. She is also completing her PhD studies at the University of Waterloo. Her research is aligned with Ad hoc Retrieval and Conversational Search in IR and NLP domains.

\noindent
\paragraph{Shrestha Mohanty} is a Machine Learning Engineer at Microsoft. She primarily works in areas of machine learning, deep learning, and natural language processing, including topics such as personalization, dialogue systems and multilingual language models. She is also interested in and has worked on problems at the intersection of machine learning and healthcare. Prior to Microsoft, she completed her master’s in information science from University of California, Berkeley.

\noindent
\paragraph{Mohammad Aliannejadi} is an assistant professor at the University of Amsterdam (The Netherlands). His research interests include single- and mixed-initiative conversational information access and recommender systems. Previously, he completed his Ph.D.~at Universit\`a della Svizzera italiana (Switzerland), where he worked on novel approaches of information access in conversations. He co-organized the Conversational AI Challenge, ConvAI 3 (EMNLP 2020).

\noindent
\paragraph{Ziming Li} is Research Scientist in Amazon Alexa, and he was previously a Postdoc at University of Amsterdam. His main research interest is developing advanced dialogue systems, including dialogue policy learning and evaluation. He is also interested in the fields of conversational search and reinforcement learning.

\noindent
\paragraph{Maartje ter Hoeve} is a PhD candidate at the University of Amsterdam. Her main research interest is how we can learn from humans and their cognition to improve our NLP and IR systems. She has a background in both Linguistics and Artificial Intelligence. She co-organized the previous IGLU challenge (NeurIPS 2021).

\noindent
\paragraph{Mikhail Burtsev}'s current research interests: application of neural nets and reinforcement learning in the NLP domain. He was a faculty advisor of MIPT team participating in Alexa Prize Challenges 3 and 4.  He proposed and organised Conversational AI Challenges: ConvAI 1 (NIPS 2017), ConvAI 2 (NeurIPS 2018), ConvAI 3 (EMNLP 2020). 

\noindent
\paragraph{Yuxuan Sun} is a Research Engineer in Facebook AI Research (FAIR). His research interests lie in natural language processing, neural symbolic learning and reasoning, and human-in-the-loop learning for embodied agents.

\noindent
\paragraph{Arthur Szlam} is a Research Scientist at Facebook AI Research.  He works on connecting perception, memory, language, and action in artificial agents.  Prior to joining Facebook, he was on the faculty of the City College of New York (CCNY), and was the recipient of a Sloan research fellowship. He has been a co-organizer of previous ConvAI challenges.

\noindent
\paragraph{Kavya Srinet} is a Research Engineering Manager at Facebook AI Research working towards a long-term goal of developing interactive assistants. She works on building assistants that can learn from interactions with humans. Prior to FAIR, she was a Machine Learning Engineer at the AI Lab at Baidu Research, where she worked on speech and NLP problems. Kavya was at the Allen Institute for Artificial Intelligence for a summer before that working on learning to rank for Semantic Scholar. Kavya did her graduate school from Language Technology Institute at Carnegie Mellon University, where she worked on areas of machine translation, question answering and learning to rank for graphs and knowledge bases.

\noindent
\paragraph{Milagro Teruel} is a Research Software Engineer at Microsoft Research. She works in developing new platforms to facilitate research and evaluation of machine learning models from human interactions in natural language. Previous to joining Microsoft, Milagro has worked in learning complex relations in text, such as argument mining on legal documents and Covid-19 diagnosis from medical histories, with focus on data collection in low-resource scenarios.

\noindent
\paragraph{Ahmed Awadallah} is a Senior Principal Research Manager at Microsoft Research where he leads the Language \& Information Technologies Group. His research has sought to understand how people interact with information and to enable machines to understand and communicate in natural language (NL) and assist with task completion. More recently, his research has focused learning form limited annotated data (e.g., few-shot learning  and transfer learning) and from user interactions (e.g. interactive semantic parsing) . Ahmed’s contributions to NLP and IR have recently been recognized with the 2020 Karen Spärck Jones Award from the British Computer Society. Ahmed regularly serves as (senior) committee, area chair, guest editor and editorial board member at many major NLP and IR conferences and journal.

\subsection{Advisory Board}

\noindent
\paragraph{Tim Rocktäschel} is an Associate Professor at the Centre for Artificial Intelligence in the Department of Computer Science at University College London (UCL), where he is leading the UCL Deciding, Acting, and Reasoning with Knowledge (DARK) Lab. He is also a Scholar of the European Laboratory for Learning and Intelligent Systems (ELLIS). He was a Manager and Research Scientist at Facebook AI Research (FAIR), a Postdoctoral Researcher in Reinforcement Learning at the Whiteson Research Lab at the University of Oxford, a Junior Research Fellow in Computer Science at Jesus College, and a Stipendiary Lecturer in Computer Science at Hertford College. Tim obtained his Ph.D. from UCL under the supervision of Sebastian Riedel, where he was awarded a Microsoft Research Ph.D. Scholarship in 2013 and a Google Ph.D. Fellowship in 2017. Tim's research focuses on Reinforcement Learning and Open-ended Learning.

\noindent
\paragraph{Julia Hockenmaier} is an associate professor at the University of Illinois at Urbana-Champaign. She has received a CAREER award for her work on CCG-based grammar induction and an IJCAI-JAIR Best Paper Prize for her work on image description. She has served as member and chair of the NAACL board, president of SIGNLL, and as program chair of CoNLL 2013 and EMNLP 2018.

\noindent

\noindent
\paragraph{Katja Hofmann} is a Senior Principal Researcher at the Machine Intelligence and Perception group at Microsoft Research Cambridge. Her research focuses on reinforcement learning with applications in video games, as she believes that games will drive a transformation of how people interact with AI technology. She is the research lead of Project Malmo, which uses the popular game Minecraft as an experimentation platform for developing intelligent technology, and has previously co-organized two competitions based on the Malmo platform. Her long-term goal is to develop AI systems that learn to collaborate with people, to empower their users and help solve complex real-world problems.

\noindent
\paragraph{Bill Dolan} is Partner Researcher Manager at Microsoft Research, where he manages the Natural Language Processing group. He has worked on a wide variety of problems, including the acquisition of structured common-sense knowledge from free text, paraphrasing, text rewriting to improve grammar and style, and most recently on data-driven, grounded approaches to handling dialog. He has helped organize a number of research community events over the years, including the RTE challenges and the ``Understanding Situated Language in Everyday Life" summer research institute with the University of Washington, as well as running the Microsoft Research graduate fellowship program from 2014-2017.

\noindent
\paragraph{Ryen W. White} is a Partner Research Area Manager at Microsoft Research, where he leads the Language and Intelligent Assistance research area. He led the applied science organization at Microsoft Cortana, and he was chief scientist at Microsoft Health. Ryen has authored hundreds of publications in areas such as information retrieval, computational health, and human-computer interaction - including many that received awards. He was program chair for SIGIR 2017 and The Web Conference 2019. Ryen is editor-in-chief of ACM Transactions on the Web.

\noindent
\paragraph{Maarten de Rijke} is Distinguished University Professor of Artificial Intelligence and Information Retrieval at the University of Amsterdam. He is VP Personalization and Relevance and Senior Research Fellow at Ahold Delhaize. His research strives to build intelligent technology to connect people to information. His team pushes the frontiers of search engines, recommender systems and conversational assistants. They also investigate the influence of the technology they develop on society. De Rijke is the director of the Innovation Center for Artificial Intelligence.

\noindent
\paragraph{Oleg Rokhlenko} is a Senior Science Manager at Amazon Alexa Shopping, leading a science team 
specializing in Natural Language Processing (NLP) and Conversational AI for voice-activated shopping assistant agents. Before joining Amazon, Oleg was a Senior Research Scientist in Yahoo Labs, focusing on ML, NLP and Community Question Answering.
Prior to that, Oleg was a Research Staff Member in IBM Research. Oleg holds a PhD from the Technion - Israel Institute of Technology and has numerous publications in the top-tier conferences in the relevant fields. Oleg also serves on the program committee of the leading ML, NLP, and IR conferences such as NIPS, ICML, ACL, EMNLP, NAACL, SIGIR, WSDM, The Web Conference, and more.

\noindent
\paragraph{Sharada Mohanty} is the CEO and Co-founder of AIcrowd, a community of AI researchers built around a platform encouraging open and reproducible artificial intelligence research. He was the co-organizer of many large-scale machine learning competitions, such as NeurIPS 2017: Learning to Run Challenge, NeurIPS 2018: AI for Prosthetics Challenge, NeurIPS 2018: Adversarial Vision Challenge, NeurIPS 2019: MineRL Competition, NeurIPS 2019: Disentanglement Challenge, NeurIPS 2019: REAL Robots Challenge, NeurIPS 2020: Flatland Competition, NeurIPS 2020: Procgen Competition. He is extremely passionate about benchmarks and building communities. During his Ph.D. at EPFL, he worked on numerous problems at the intersection of AI and health, with a strong interest in reinforcement learning. In his previous roles, he has worked at the Theoretical Physics department at CERN on crowdsourcing compute for PYTHIA powered Monte-Carlo simulations; he has had a brief stint at UNOSAT building GeoTag-X---a platform for crowdsourcing analysis of media coming out of disasters to assist in disaster relief efforts. In his current role, he focuses on building better engineering tools for AI researchers and making research in AI accessible to a larger community of engineers.

%% file: 04-appendix.tex
\newpage
\section*{Appendix}

\begin{figure}[ht!]
\centering
    \includegraphics[ width=\textwidth]{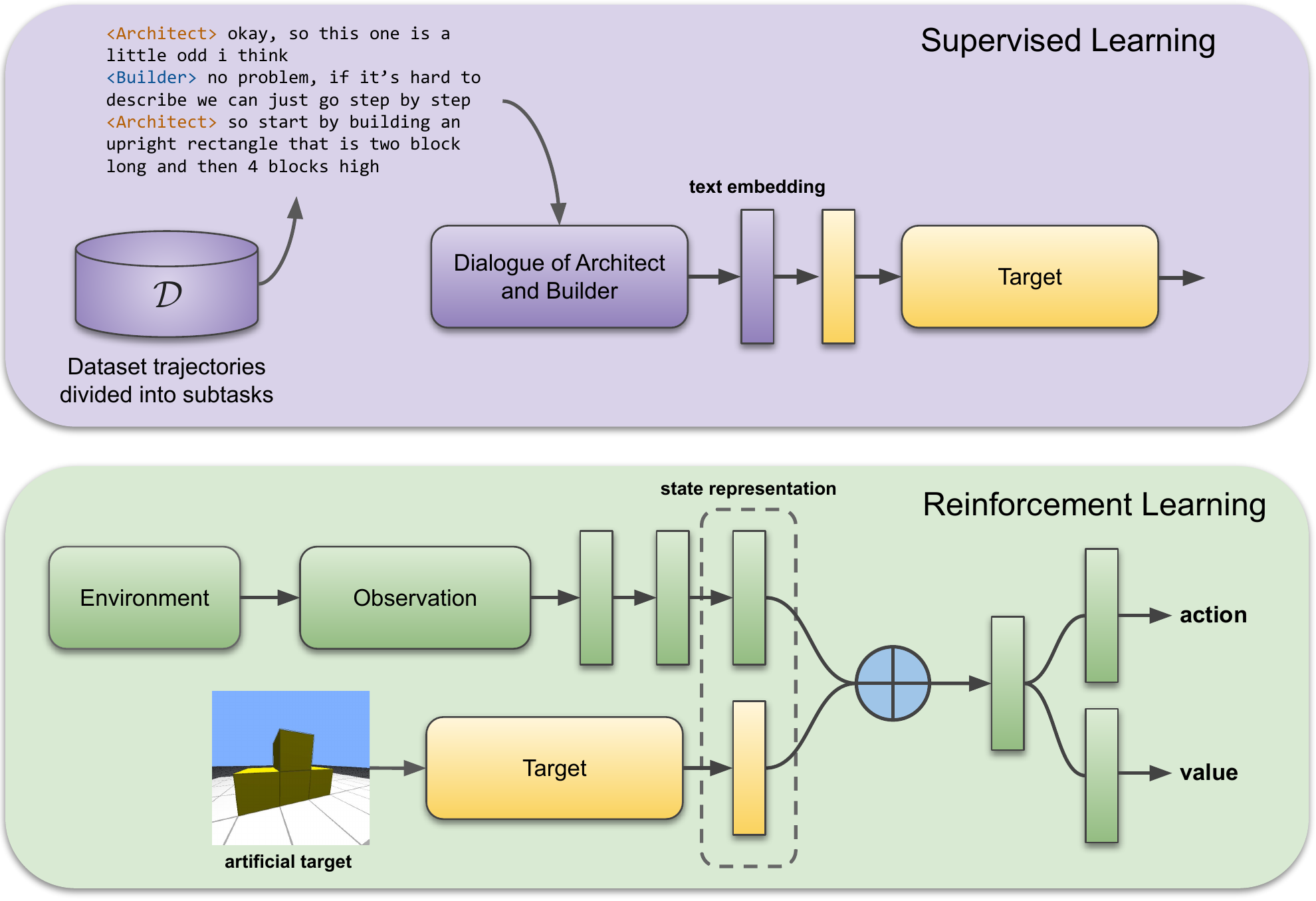}
   \caption{The decomposition scheme of silent builder. There two main parts, which should be solved separately using NLU and RL methods.}
   \vspace{-0.5cm}
   \label{fig:baseline-decomposition}
\end{figure}

\section{Maximal Intersection Calculation Scheme}
\label{reward_calc}

We calculate reward using the following algorithm. Effectively, at each step, we calculate the maximal intersection between the current and the target grids as described in Algorithm \ref{max_int_alg}. The reward is then defined in terms of maximal intersections at two consecutive steps, that is

\begin{gather*}
    r_t \leftarrow \texttt{max\_int}_{t} - \texttt{max\_int}_{t-1}.
\end{gather*}

Algorithm \ref{max_int_alg} shows a principle version of the intersection calculation. In practice, we optimize it as takes a lot of time to compute it using the naive version. First, we recalculate the intersection size only when we observe a change in the building zone (i.e., one block added or removed). Second, there can be some translation configuration such that the full intersection is not possible (e.g., for a goal structure with block in the four corners of the building zone, if the agent has placed a block in the center of the zone, then this means it would be unable to have full intersection since any non-zero shift would cut out at least one of these blocks). Therefore, we consider only those intersections that do not cut the target grid.

\begin{algorithm*}[t!]
\label{max_int_alg}
\caption{Maximal Intersection Pseudocode}
\SetKwInOut{KwIn}{Input}
\SetKwInOut{KwOut}{Output}
\KwIn{\texttt{current}: Current build zone snaphot, 3d tensor $(y, x, z)$ filled with integers in range $[0, 6]$; \\
\texttt{target}: Target build zone, 3d tensor $(y, x, z)$ filled with integers in range $[0, 6]$; \\
\texttt{X, Y, Z}: sizes of the build zone over three axes
}

\KwOut{\texttt{max\_int}: Scalar intersection size}
\texttt{maximum} $\leftarrow 0$;\\ 
\tcp{\texttt{Iterate over rotations}}
\For{$i \in [1 .. 4]$}{
    \tcp{\texttt{Iterate over translations}}
    \For{$dx \in [-\texttt{X} .. \texttt{X}$]}{
        \For{$dz \in [-\texttt{Z} .. \texttt{Z}$]}{
            \texttt{shifted\_grid} $\leftarrow translate(\texttt{current}, dx, dz);$\\
            \texttt{intersection} $\leftarrow (\texttt{shifted\_grid} = \texttt{target}) \land (\texttt{target} \neq 0)$;\\
            \texttt{maximum} $\leftarrow \max(\texttt{maximum}, \texttt{intersection})$;\\
        }
    }
    \texttt{current} $\leftarrow rotate\_90degrees(\texttt{current})$;\\
}
\Return{\texttt{maximum}}
\end{algorithm*}

\section{Competition Rules}
\subsection{Silent Builder}
\label{b_rules}

\begin{enumerate}
    \item Submissions to the IGLU competition must be open. We expect Teams to reveal details of their method, including source-code and machine learning model parameters.
    \item A team can submit separate entries to both (Clarifying Questions and Builder) tracks; performance in the tracks will be evaluated separately. Submissions between the phases are not linked in any way.
    \item Interaction with the submission system must be through the provided interface, in particular through the “step” function. Only the iglu gym environment provided by organizers can be used. Additional information may not be extracted from the simulator in any way (unless the opposite stated for a specific phase).
    \item Submissions will be evaluated in IGLUSilentBuilderVisual-v0 or IGLUSilentBuilder-v0 depending on a phase.
    \item In IGLUSilentBuilderVisual-v0 only image, chat, compass, and inventory are exposed. The use of any other fields or other environment info is prohibited during the evaluation.
    \item In IGLUSilentBuilder-v0 the allowed observation components are image, chat, compass, inventory, agent position, current grid
    \item For training machine learning models, any information, including the agent position and the grid state can be used.
    \item The trained machine learning models can use either a public dataset from the Minecraft Dialogue or public IGLU dataset or both of them. If any other sources of data are used, they should also be publicly available.
    \item The submission to the Builder competition can use a trained machine learning model or the hardcoded heuristic-based or rule-based agents. 
    \item Participants do not have the access to the test examples which their agents are evaluated on. Any unauthorized access to the information that can reveal the test tasks contents will lead to disqualification.  
    \item The use of internet will be turned off for a submission to avoid cheating. Every downloadable data (e.g., packages,datasets,neural networks weights) should be included in submission or installed in the image.
\end{enumerate}

\section{Preliminary Set of Skills in the Builder Track}
\label{skills}

A preliminary set of agent skills which are needed to build a structure is provided below. Note that these labels do not correspond to structures themselves but to a behavior (and an instruction) that led to a particular structure.

\begin{itemize}
    \item \textbf{flat:} flat structure with all blocks on the ground
    \item \textbf{flying:} there are blocks that cannot be placed without removing some other blocks (i.e. one block or a group of blocks are not connected to the ground)
    \item \textbf{tricky:} some blocks are hidden or there should be a specific order in which they should be placed (e.g., a 27 blocks cube with the block in the center of different color than the whole cube)
    \item \textbf{tall:} a structure cannot be built without the agent being at a sufficient height (the placement radius is 3 blocks)
\end{itemize}

Each submission is evaluated on a set of tasks with certain skill labels associated with each task. The platform will then respond with a per skill performance that could help participants to better understand the performance of their submission.

\section{Example of The Behavior Data}
\label{behavior_data}

In this section we show an example of the raw behavioral data represented as an event stream in the JSON format.

\begin{lstlisting}[language=json]
{'gameId': 19,
 'stepId': 1,
 'avatarInfo': {'pos': [0.16371048991651502,
   66.19999999999993,
   5.601672092120345],
  'look': [-0.6139999999999998, 0.001999999999999995]},
 'worldEndingState': {'blocks': [[0, 64, 1, 50],
   [1, 64, 1, 50],
   [3, 63, 1, 59],
   [3, 65, 1, 59],
   [3, 66, 1, 59],
   [4, 63, 1, 59],
   [4, 65, 1, 59],
   [5, 63, 1, 59],
   [5, 64, 1, 59],
   [5, 65, 1, 59],
   [5, 66, 1, 59]]},
 'clarification_question': 'null',
 'tape': '
   action start_recover_world_state
   action select_and_place_block 59 3 63 1 6.93311074709203e-24  64.12000274658203 4 -1.733277824122399e-24 0 -1
   action select_and_place_block 59 3 65 1 6.93311074709203e-24 64.12000274658203 4 -1.733277824122399e-24 0 -1
     ...
   action select_and_place_block 59 5 66 1 6.93311074709203e-24 64.12000274658203 4 -1.733277824122399e-24 0 -1
   block_change  (3, 63, 1, 0, 59) (3, 65, 1, 0, 59) (3, 66, 1, 0, 59) (4, 63, 1, 0, 59) (4, 65, 1, 0, 59) (5, 63, 1, 0, 59) (5, 64, 1, 0, 59) (5, 65, 1, 0, 59) (5, 66, 1, 0, 59)
   pos_change (0.7800644185765344, 66.83999999999992, 3.38537289325922)
   set_look (-0.4079999999999997, -0.3500000000000002)
   action finish_recover_world_state
     ...
   action step_backward
   pos_change (0.780392971263885, 66.83999999999992, 3.4675106241016485)
   set_look (-0.756, 0.00599999999999988)
   pos_change (0.7812531261284168, 66.83999999999992, 3.610868043629237)
   set_look (-0.758, 0.00599999999999988)
     ...
   action step_right
   pos_change (0.8616502838004294, 66.83999999999992, 5.380599639872034)
   set_look (-0.758, -0.032000000000000126)
   pos_change (1.004321268291321, 66.83999999999992, 5.4044179797943865)
   set_look (-0.754, -0.032000000000000126)
     ...
   pos_change (1.861788332305804, 65.39999999999995, 4.277191269822069)
   action select_and_place_block 50 1 64 1 2.2324962615966797 69.09007263183594 7.2514328956604 -0.09427810765305872 -0.6472718661639695 -0.7564064844314669
   block_change  (1, 64, 1, 0, 50)
   pos_change (1.8616574997357698, 65.39999999999995, 4.276141581759024)
   pos_change (1.8615528336797424, 65.39999999999995, 4.275301831308588)
   ...
   '
}
\end{lstlisting}